\documentclass[runningheads]{llncs}
\usepackage[T1]{fontenc}
\usepackage{indentfirst}
\usepackage{graphicx}
\usepackage{times}
\usepackage{helvet}
\usepackage{courier}
\usepackage{amsmath}
\usepackage{algorithm}
\usepackage{algorithmic}
\usepackage{csquotes} 
\usepackage{color}
\usepackage{paralist}
\usepackage{amssymb}
\usepackage{indentfirst}
\usepackage{subfigure}
\usepackage{float}
\usepackage{multirow}
\usepackage{cite}
\usepackage{mathrsfs}

\usepackage{mathrsfs} 
\usepackage{amsfonts}
\usepackage{todonotes}
\usepackage{pgfplots} 
\usepackage{epstopdf}
\usepackage{epsfig}
\pgfplotsset{compat=newest}
\usepackage{color}
\definecolor{forestgreen}{RGB}{0,139,69}
\usepackage{caption}
\usepackage{tabularx}
\usepackage[colorlinks, linkcolor=red,  anchorcolor=blue, citecolor=blue]{hyperref}

\begin{document}
\title{
Exploiting Memory-aware Q-distribution Prediction for Nuclear Fusion via Modern Hopfield Network
\thanks{Corresponding author: Xiao Wang (\email{xiaowang@ahu.edu.cn})}  
}

%


\author{
Qingchuan Ma \inst{1} \and
Shiao Wang \inst{1} \and 
Tong Zheng \inst{1} \and 
Xiaodong Dai \inst{1} \and \\ 
Yifeng Wang \inst{2} \and 
Qingquan Yang \inst{2} \and 
Xiao Wang \inst{1} 
}

\authorrunning{Qingchuan Ma et al.}
%
\institute{
School of Computer Science and Technology, Anhui University, Hefei, China \and 
Institute of Plasma Physics, Chinese Academy of Sciences, Hefei, China 
}


%
\maketitle              
\begin{abstract}
This study addresses the critical challenge of predicting the Q-distribution in long-term stable nuclear fusion task, a key component for advancing clean energy solutions. We introduce an innovative deep learning framework that employs Modern Hopfield Networks to incorporate associative memory from historical shots. Utilizing a newly compiled dataset, we demonstrate the effectiveness of our approach in enhancing Q-distribution prediction. The proposed method represents a significant advancement by leveraging historical memory information for the first time in this context, showcasing improved prediction accuracy and contributing to the optimization of nuclear fusion research.

\end{abstract}

\section{Introduction}

Controlled nuclear fusion is one of the key approaches expected to solve humanity's challenge of clean energy. Currently, the main international and domestic nuclear fusion devices include ITER, EAST, HL-2M, HT-7, etc. Research in the field of nuclear fusion, powered by artificial intelligence, is beginning to attract the attention of an increasing number of researchers. This includes areas such as disruption prediction~\cite{kates2019predicting,churchill2020deep}, deep reinforcement learning based control system optimization~\cite{seo2021feedforward}, magnetic control method~\cite{degrave2022magnetic}, and more.

In long-term stable nuclear fusion missions, the prediction of the Q-distribution is one of the key points, and this article explores this issue. As it is a new research problem for Q-distribution prediction using artificial intelligence, the collection of training and testing data is the first thing we need to do. Therefore, we used a simulation program to generate input signals containing various physical quantities, along with the corresponding Q-distribution ground truth. Finally, we obtain a dataset containing 5753 paired input-output samples and split them into a training and testing subset which contains 5166 and 587 samples, respectively.

Based on the newly collected database, in this paper, we propose to predict the Q-distribution using deep neural networks. Considering the raw data of each physical variable are all 1-D vectors in a sample, it is intuitive to encode them using an MLP (Multi-Layer Perceptron) network and regress the Q-distribution using a dense layer. It will be a simple baseline but not a good approach for high-performance Q-distribution prediction. Because the samples collected from nearby shots are highly correlated to each other, the performance will be better once the history memory information is aggregated. As shown in Fig.~\ref{fig:framework}, in this work, we propose to encode associative memory of context shots using the Modern Hopfield Networks (MHN). Specifically, we take the historical example as another input and extract their features using the shared MLP. The two input features will be concatenated and added with position encoding features, then, we feed them into the MHN for associative memory augmented feature representation learning. Meanwhile, we also concatenate the learnable parameters with features of raw signal and feed them into the MHN. These two features will be concatenated and fed into the dense layer for Q-distribution prediction. Extensive experiments demonstrate the effectiveness of our proposed memory-augmented network for Q-distribution prediction.

To sum up, we draw the contributions of this work as the following two aspects: 

1). We propose a novel associative memory-augmented Q-distribution prediction framework using modern Hopfield networks. It is the first time to exploit the historical memory information for this research task. 

2). We conducted extensive experiments on the newly collected dataset, and the results demonstrate the effectiveness of historical memory.

\section{Our Proposed Approach} \label{sec:method}


\subsection{Problem Formulation} \label{sec:probFormulation} 
A key challenge in nuclear fusion research is the accurate prediction of the Q-distribution, which plays a pivotal role in determining plasma stability and behavior within a tokamak. However, the highly nonlinear and complex dynamics of plasma physics make this task exceedingly difficult for traditional empirical models, which are typically based on physics-driven insights. We formulate this problem as a regression task~\cite{lathuiliere2019comprehensive}, where the model takes time-series data as input and outputs the Q-distribution. To address the limitations of conventional methods, we propose a novel deep learning approach leveraging the Modern Hopfield Network for history memory-guided learning. By utilizing the memory mechanisms of the Hopfield Network, we provide a more reliable and effective solution for Q-distribution prediction.

\subsection{Overview}

\begin{figure*}
\centering
\includegraphics[width=\textwidth]{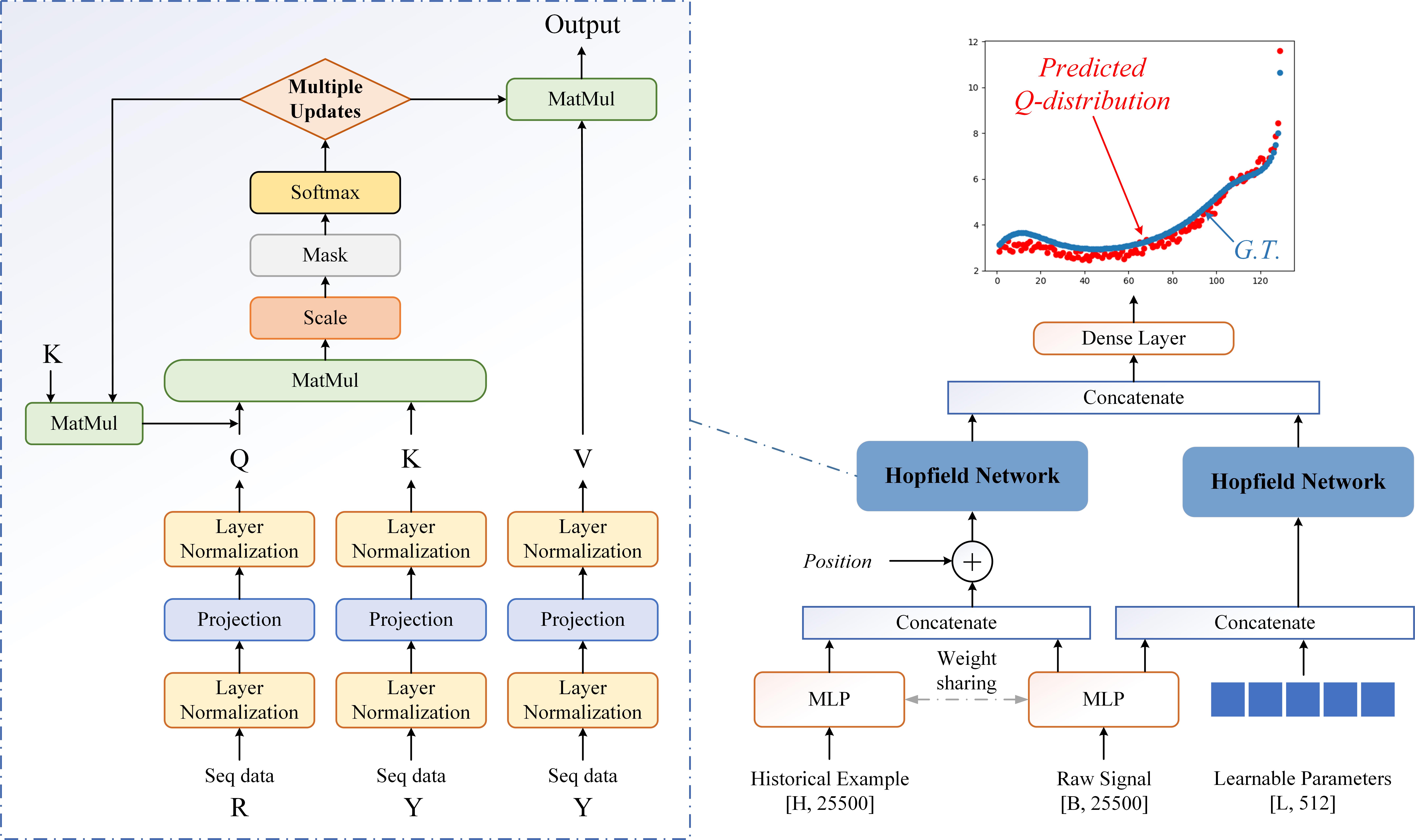}
\caption{An overview of our proposed associative memory augmented Q-distribution using the Modern Hopfield Networks.}  
\label{fig:framework}
\end{figure*}

In this paper, we delve into the practical significance of utilizing the memory storage capabilities of the Hopfield Network in predicting nuclear fusion sequence information within the realm of physics applications. To maximize the accuracy of predictive performance, our focus lies in treating nuclear fusion data as sequential data with temporal information and utilizing the Hopfield Network to extract temporal sequence features. Specifically, within each shot, there are multiple temporal samples. We first identify historical samples for each sample and utilize the Hopfield Network to assist in extracting historical information for the current sample. It has been proven that this approach of historical feature extraction enhances the predictive capability of Q-distribution. By incorporating historical information extracted by the Hopfield Network and global information into the features of current samples, we strengthen the overall predictive capability. Our network framework is illustrated in Figure~\ref{fig:framework}.


\subsection{Input Processing}  
In this study, our original nuclear fusion data comprises a total of 141 indicators, with each indicator consisting of multiple floating-point numbers. The $npsip-q$ and $q$ indicators represent the Q-distribution that we aim to predict (i.e., the model output labels), while all other indicators represent information regarding the state of plasma disruption during the nuclear fusion process. We concatenate the other 139 indicators together as a vector, which serves as the input. For each sample, the input consists of a total of 25,500 dimensions. For a batch, the input shape is (batch size, 25,500), and the label shape is (batch size, 129). Through this approach, we ensure that each batch has both input and label data~\cite{TAWAKULI2024}.

Additionally, the nuclear fusion data consists of 22 shots, with each shot comprising over 200 different samples at different times. Therefore, the nuclear fusion data can be viewed as sequential data, where the data within each shot is sorted according to time, resulting in a sequence of data for each shot. Consequently, we treat each shot as a sequence. For instance, considering a particular sample as the current sample, the samples preceding it in time constitute the historical samples of this sample. This enables us to obtain the historical samples for each sample, which are then inputted into the model. If the time gap between samples is too large, the historical information is disregarded. Furthermore, for testing samples to extract historical features effectively, they also require historical samples. Hence, during the process of extracting the test set, continuous samples are extracted to ensure that the vast majority of samples have historical information from their adjacent samples.


\subsection{Network Architecture} 
Based on the concept of deep learning for sequential data and global features, we have designed a network model capable of predicting Q-distribution in nuclear fusion tasks, as depicted in Figure~\ref{fig:framework}. The main components of the model include two Hopfield Network modules, which are utilized for extracting features from historical data and global features, respectively, aiding us in prediction. The detailed process overview is outlined as follows.

\noindent $\bullet$ \textbf{Raw Data Processing.~} 
By processing the data labeled with 139 indicators of nuclear fusion, we can transform the raw data into tensor form to be used as input, with the input size expressed as $\mathbb{R}^{B \times D}$. Here, $B$ represents the batch size, indicating the number of samples included in each batch during training or testing, while $D$ denotes the number of features for each individual data point, also referred to as the data dimension, with a specific value of 25500. We feed a batch of data into the neural network, as depicted in Figure~\ref{fig:framework}. We utilize an MLP as the basic feature extractor, extracting original features from the raw data. MLP, or Multi-Layer Perceptron, comprises multiple linear layers and multiple activation functions~\cite{dubey2022activation}, where the activation functions introduce nonlinear features. This can be summarized as follows:
\begin{equation}
    \label{mlp_equation} 
feature = Linear((\sigma(Linear(V_i)))*N)
\end{equation}
where $N$ represents the number of such combinations of linear layers and activation functions. The features extracted here only represent the characteristics of individual samples and require further processing to enable learning of historical and global features. The historical samples can also be inputted into the MLP to obtain features for the historical samples.

\noindent $\bullet$ \textbf{Incorporating Temporal Information.~}
Historical information and current sample features are concatenated along a new dimension, commonly referred to as the sequence length in deep learning. This yields new features represented as $\mathbb{R}^{B \times S \times D}$, where $S$ denotes the quantity of historical information plus the number of current samples. After processing with the position embedding module, positional information, or in other words, temporal information, can be incorporated. The position embedding module is used to add positional encoding to sequential data. Here, we employ sine and cosine functions to compute the positional encoding~\cite{wolf-etal-2020-transformers}, which can be expressed by the following equation:
\begin{equation}
    \label{PE_i} 
    PE_{(pos, 2i)} = \sin\left(\frac{pos}{10000^{\frac{2i}{\text{hidden size}}}}\right)
\end{equation}
\begin{equation}
    \label{PE_2i+1} 
    PE_{(pos, 2i + 1)} = \cos\left(\frac{pos}{10000^{\frac{2i}{\text{hidden size}}}}\right)
\end{equation}
\begin{equation}
    \label{PE} 
    x_{(b, s, h)} = x_{(b, s, h)} + PE_{(1, s, h)}
\end{equation}
At this point, the features represent true sequential data, enriched with temporal information.

\noindent $\bullet$ \textbf{Extraction of Historical and Global Features.~}
The Hopfield Network is employed to further process features, serving as a module equipped with memory storage capabilities. When sequential features are inputted into the Hopfield Network, it can extract historical information features based on the similarity between the current sample and historical sample features. The output shape remains the same as $\mathbb{R}^{B \times S \times D}$, and at this point, the features of the last sequence (the most recent features in the temporal dimension), which represent the features of the current sample, are extracted. In comparison to features extracted by MLP, these features contain historical information, thereby completing the feature extraction from historical information. The process of extracting global features is similar. The detailed architecture is depicted in Figure~\ref{fig:framework}. The processing formula of the Hopfield Network can be summarized as follows:
\begin{equation}
    \label{Hopfield} 
Hopfield(R, Y, Y) = softmax(\beta\cdot R W_Q W_K^t Y^t)Y W_K W_v
\end{equation}
Here is a detailed explanation of the calculation process of the Hopfield Network. $\beta$ is a hyperparameter. All $W$ matrices represent projections of dimensions, implemented using linear layers. For instance, $W_R$ is used to transform the feature dimension (the last dimension) to the Hopfield space of dimension $D$. Similarly, $W_K$ and $W_V$ serve the same purpose. With these projections, we can ignore the dimension mapping, thus simplifying the formula to the following:
\begin{equation}
    \label{Hopfield_s} 
Hopfield(R, Y, Y) = softmax(\beta\cdot R Y^t)Y
\end{equation}
where $R$ represents the state (query) patterns to be retrieved, while $Y$ represents the stored key patterns. Assuming both $R$ and $Y$ have shapes of ($B$, $S$, $D$), first, matrix operations are performed between $R$ and $Y$ along the $D$ dimension, enabling the calculation of the similarity between a certain pattern and other patterns. The resulting shape is ($B$, $S$, $S$). For example, $r_{b,{i},{j}}$ represents the similarity between the $i^{th}$ state pattern of $R$ and the $j^{th}$ stored pattern of $Y$ in the $b^{th}$ batch.

Then, the obtained similarity matrix is multiplied with $Y$ along the $S$ dimension, allowing for pattern retrieval from $Y$ based on similarity. The resulting shape is ($B$, $S$, $D$). It represents the patterns retrieved from $Y$ using $R$ as the query. If both $R$ and $Y$ are regarded as sequence features obtained through position embedding, this process can be viewed as using the Hopfield Network to extract information from different positions in the sequence features, thereby extracting historical information. The features of the last pattern outputted by the Hopfield Network correspond to the features of the current sample with historical information incorporated. The method for extracting global features is similar to that for extracting historical features. Finally, the output features of the two Hopfield Networks are concatenated and then inputted into a linear layer to project the features into the output dimension, yielding the model's prediction results.

\subsection{Loss Function}  
Predictions can be obtained by passing the data through the final linear layer of the model. We utilize the Mean Squared Error ($MSE$) loss function~\cite{yessou2020comparative,mathieu2015deep} to gauge the disparity between the predicted results and the ground truth (the actual Q-distribution). The specific computation process of the loss function can be expressed by Equation~\ref{MSEloss}. 
\begin{equation}
    \label{MSEloss} 
MSE = \frac{1}{n} \sum_{i=1}^{n} (GT_i - Q_i)^2
\end{equation}
In the equation, $Q_i$ represents the distribution prediction result for the $x_i$ sample generated by the model, while $GT_i$ denotes the Q-distribution of the $x_i$ sample.

\section{Experiments} \label{sec:experiments}

\subsection{Dataset and Evaluation Metric} 
In the nuclear fusion dataset, comprising 22 shots, a total of 5753 samples were recorded. The temporal resolution between adjacent samples is notably fine, approximately 10 milliseconds. To construct our test set, we adopted a consistent ratio (1/10) for extracting continuous samples from each shot. It is worth noting that the vast majority of samples have adjacent historical samples in time. Ultimately, we partition the nuclear fusion dataset into training and testing subsets, comprising 5166 and 587 raw data points, respectively. To evaluate our model's performance, we utilize the $MSE$ (Mean Squared Error) metric, which quantifies the disparity between our predicted Q-distribution and the ground truth.

\subsection{Implementation Details} 
The models were trained end-to-end. All experimental results were obtained after 140 epochs of training. A learning rate of 0.001 and a batch size of 16 were utilized, and training was performed using the $SGD$ optimizer~\cite{sun2019survey}. Subsequently, the models were evaluated on the test set using the $MSE$ loss function to obtain the error results. The code was implemented in Python using the PyTorch~\cite{paszke2019pytorch} framework. The computations were carried out on a server equipped with a CPU Intel(R) Xeon(R) CPU E5-2620 v4 @ 2.10GHz and a GPU TITAN XP with 12 GB of memory. 

\begin{table}
\center
\caption{Comparison with Other Models.} 
\label{comparison}
\begin{tabular}{c|cccccc}
\hline 
\textbf{Algorithm}  & \textbf{MLP} &\textbf{RNN} &\textbf{GRU} &\textbf{LSTM} &\textbf{Transformer} &\textbf{Ours}\\
\hline 
\textbf{MSE}        &0.0666              &0.0656       &0.0670       & 0.0700       &0.0659               &0.0584\\
\hline 
\end{tabular}
\end{table}

\subsection{Comparison with Other Algorithms}  
To effectively extract visual features from sequential data, we compared various mainstream sequence models~\cite{lim2021time} to determine the optimal approach. Table~\ref{comparison} summarizes our findings. We evaluated several models, including RNNs, GRUs, LSTMs, Transformers, and our Hopfield Network. Through comprehensive experimental comparisons, we observed that employing the Hopfield Net as our visual backbone network yielded the best performance. This can be attributed to the intrinsic memory storage capability and powerful feature extraction ability of the Hopfield Network, which helps extract historical information from sequential data. Surprisingly, the main computational formulas of the Transformer and Hopfield Network are quite similar, but there is a significant performance gap. We speculate that this is because the Transformer (specifically, the encoder layer of the Transformer model we used) is more complex, and our dataset is not large enough, which can lead to overfitting. In contrast, our Hopfield Network is more flexible to use and has been optimized specifically for the nuclear fusion task, allowing it to demonstrate the best results.

\subsection{Ablation Study}  

\begin{table}
\center
\caption{Component analysis results. LParam is short for Learnable Parameter.} 
\label{CAResults} 
\begin{tabular}{c|c|c|c|c|c}  
\hline 
\textbf{No.}  & \textbf{MLP}  &\textbf{Hopfield}  &\textbf{Position} &\textbf{LParam} &\textbf{MSE}   \\
\hline 
1 &\checkmark   &              &             &             &0.0666      \\
2 &\checkmark   &\checkmark    &             &             &0.0620      \\
3 &\checkmark   &\checkmark    &\checkmark   &             &0.0593      \\
4 &\checkmark   &              &             &\checkmark   &0.0658      \\
5 &\checkmark   &\checkmark    &\checkmark   &\checkmark   &0.0584      \\
\hline
\end{tabular} 
\end{table}

\begin{figure}[htbp]
    \centering
    \small
    \begin{tikzpicture}
        \begin{axis}[
            ybar,
            symbolic x coords={256, 512, 1024, 2048, 4096},
            xtick=data,
            ylabel={MSE},
            ymin=0.05,
            ymax=0.07,
            bar width=15pt,
            width=0.8\textwidth,
            height=0.5\textwidth,
            nodes near coords,
            every node near coord/.style={font=\small, yshift=3pt, /pgf/number format/.cd, fixed, fixed zerofill, precision=4}, 
            enlarge x limits=0.15,
            xlabel={Hidden Size},
            grid=major,
            ymajorgrids=true,
            xmajorgrids=true,
            ytick={0.05, 0.055, 0.06, 0.065, 0.07, 0.075, 0.08},
            yticklabels={0.050, 0.055, 0.060, 0.065, 0.070, 0.075, 0.08}, 
            scaled y ticks=false 
        ]
        \addplot[fill={cyan!70}] coordinates {(256,0.0636) (512,0.0654) (1024,0.0643) (2048,0.0584) (4096,0.0681)};
        \end{axis}
    \end{tikzpicture}
    \caption{Impact of Hidden Size on Hopfield Network for Historical Samples}
    \label{fig:H_hidden_size}
\end{figure}
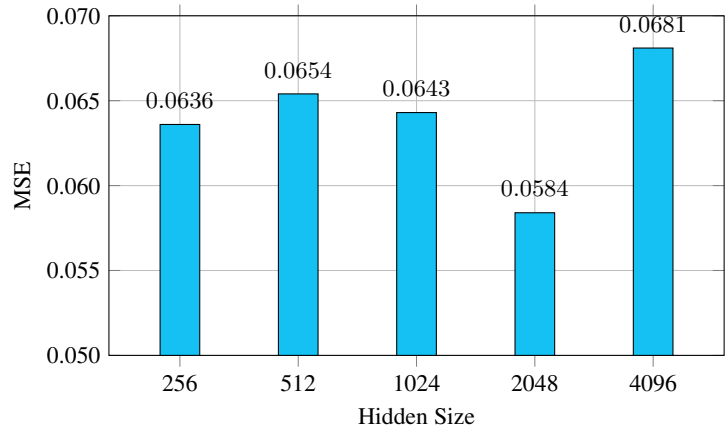

\noindent $\bullet$ \textbf{Component Analysis.~} 
This paper explores the application of deep learning in predicting nuclear fusion results. Our experiments demonstrate that treating nuclear fusion data as sequential information and utilizing the memory storage capabilities of the Hopfield Network can effectively reduce the fitting error in the Q-distribution prediction task. Initially, when the data undergoes processing through several linear layers, such as Multi-Layer Perceptron (MLP), although achieving fitting, the results are unsatisfactory with an $MSE$ loss of 0.0666, as indicated in Table~\ref{CAResults}. However, when we leverage the Hopfield Network effectively functions as an attention mechanism, improving feature extraction capability, and reducing the $MSE$ loss to 0.0620. Remarkably, introducing temporal information via the position embedding module, which provides positional information for samples at different times, significantly enhances fitting performance, resulting in an $MSE$ loss of 0.0593. Furthermore, the incorporation of a module with learnable parameters aids in extracting global information, further enhancing the predictive performance of Q-distribution, leading to a decrease in the $MSE$ loss to 0.0584. For comparison, experiments using only MLP and the module with learnable parameters were conducted. It is evident that compared to the effectiveness of the Hopfield Network and position embedding module, the module with learnable parameters does not exhibit such prominent performance. These experimental results emphasize the effectiveness of treating nuclear fusion data as sequential data and utilizing the Hopfield Network with position embedding, both of which significantly enhance the predictive performance of the Q-distribution. Moreover, incorporating a module with learnable parameters allows for learning global information, further improving the effectiveness of the model.

\begin{figure}[htbp]
    \centering
    \small
    \begin{tikzpicture}
        \begin{axis}[
            symbolic x coords={1, 2, 4, 8},
            ybar,
            xtick=data,
            ylabel={MSE},
            ymin=0.05,
            ymax=0.07,
            bar width=15pt,
            width=0.8\textwidth,
            height=0.5\textwidth,
            nodes near coords,
            every node near coord/.style={font=\small, yshift=3pt, /pgf/number format/.cd, fixed, fixed zerofill, precision=4}, 
            enlarge x limits=0.15,
            xlabel={Number of Heads},
            grid=major,
            ymajorgrids=true,
            xmajorgrids=true,
            ytick={0.05, 0.055, 0.06, 0.065, 0.07, 0.075, 0.08},
            yticklabels={0.050, 0.055, 0.060, 0.065, 0.070, 0.075, 0.08}, 
            scaled y ticks=false 
        ]
        \addplot[fill={cyan!50}] coordinates {(1,0.0672) (2,0.0584) (4,0.0586) (8,0.0610)};
        \end{axis}
    \end{tikzpicture}
    \caption{Ablation studies on the number of heads of MHN for historical samples.}
    \label{fig:H_head_num}
\end{figure}
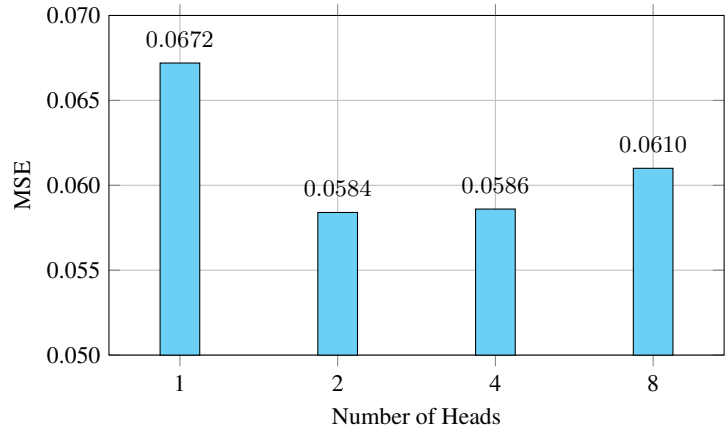

\noindent $\bullet$  \textbf{Analysis of the Hidden Size of Hopfield Network for Historical Samples.~} 
The hidden size of the Hopfield Network is also an important factor affecting the results. A larger hidden size introduces more learnable parameters, but it also poses a risk of overfitting. Thus, we set the Hopfield Network with different hidden sizes to find the most suitable hidden size, and the experimental results are shown in Fig.~\ref{fig:H_hidden_size}. We can observe that the best result is achieved when we choose a hidden size of 2048 for input, with an error of 0.0584.

\noindent $\bullet$ \textbf{Analysis of the Number of Heads of Hopfield Network for Historical Samples.~} 
Like the multi-head attention mechanism in Transformers, increasing the number of heads in the Hopfield Network allows it to extract different features. These features can then be mixed through an MLP, resulting in more powerful feature extraction. Therefore, the number of heads is also a significant parameter for enhancing the effectiveness of the Hopfield Network model. As shown in Fig.~\ref{fig:H_head_num}, we obtained different $MSE$ loss results by changing the hyperparameter, the number of heads. Here, the minimum $MSE$ loss is achieved when the number of heads is set to 2.

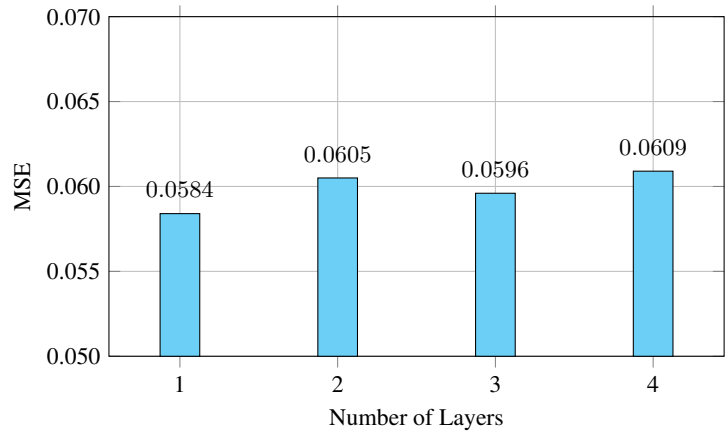
\begin{figure}[htbp]
    \centering
    \small
    \begin{tikzpicture}
        \begin{axis}[
            symbolic x coords={1, 2, 3, 4},
            ybar,
            xtick=data,
            ylabel={MSE},
            ymin=0.05,
            ymax=0.07,
            bar width=15pt,
            width=0.8\textwidth,
            height=0.5\textwidth,
            nodes near coords,
            every node near coord/.style={font=\small, yshift=3pt, /pgf/number format/.cd, fixed, fixed zerofill, precision=4}, 
            enlarge x limits=0.15,
            xlabel={Number of Layers},
            grid=major,
            ymajorgrids=true,
            xmajorgrids=true,
            ytick={0.05, 0.055, 0.06, 0.065, 0.07, 0.075, 0.08},
            yticklabels={0.050, 0.055, 0.060, 0.065, 0.070, 0.075, 0.08}, 
            scaled y ticks=false 
        ]
        \addplot[fill={cyan!50}] coordinates {(1,0.0584) (2,0.0605) (3,0.0596) (4,0.0609)};
        \end{axis}
    \end{tikzpicture}
    \caption{Ablation studies on the number of layers of Hopfield Network for historical samples.}
    \label{fig:H_layer_num}
\end{figure}




\noindent $\bullet$ \textbf{Analysis of the Number of Layers of Hopfield Network for Historical Samples.~} 
A Hopfield Network with more layers tends to have stronger fitting capabilities but also comes with fitting risks. Conversely, a Hopfield Network with fewer layers exhibits weaker fitting capabilities and may lead to underfitting issues. Therefore, finding the appropriate number of layers is also an important hyperparameter issue. We experimented with different numbers of layers and recorded the results in Fig.~\ref{fig:H_layer_num}. We found that setting the number of layers to 1 achieved the best performance.

\section{Conclusion} 
This study has successfully developed a novel approach to predicting the Q-distribution in long-term stable nuclear fusion missions by leveraging deep neural networks and associative memory augmentation. The collection and utilization of a comprehensive dataset comprising 5753 paired input-output samples have provided a solid foundation for our research. By introducing Modern Hopfield Networks (MHN) to encode associative memory of context shots, we have enhanced the feature representation learning for Q-distribution prediction. The proposed framework not only incorporates historical memory information for the first time in this context but also demonstrates significant improvements in prediction accuracy through extensive experimentation.


%
%
%
%

\bibliographystyle{splncs04}
\bibliography{reference}
\end{document}